\documentclass[10pt,journal]{IEEEtran}

\usepackage{cite}
\usepackage{amsmath,amssymb,amsfonts}
\usepackage{algorithmic}
\usepackage{graphicx}
\usepackage{textcomp}
\usepackage{booktabs}
\usepackage{algorithm}
\usepackage{multirow}
\usepackage{url}
\usepackage{bm}

\begin{document}

\title{Comparative Analysis of 3D Convolutional and 2.5D Slice-Conditioned
U-Net Architectures for MRI Super-Resolution via Elucidated Diffusion Models}

\author{Hendrik~Chiche, Eshaan~Gulati, Ludovic~Corcos, and Logan~Rouge%
\thanks{H.~Chiche is with the University of California, Berkeley, Berkeley, CA 94720 USA
(e-mail: hendrik\_chiche@berkeley.edu).}%
\thanks{E.~Gulati is with Purdue University, West Lafayette, IN 47907 USA
(e-mail: gulatie@purdue.edu).}%
\thanks{L.~Corcos and L.~Rouge are with Universit\'{e} Clermont Auvergne, Clermont-Ferrand, France
(e-mail: ludovic.corcos@etu.uca.fr; logan.rouge@etu.uca.fr).}%
\thanks{This work was supported in part by GENCI (Grand \'{E}quipement National de Calcul Intensif)
for access to high-performance computing resources. Corresponding author: Hendrik Chiche.}%
}

\markboth
{Chiche \MakeLowercase{\textit{et al.}}: 3D vs.\ 2.5D U-Net for MRI Super-Resolution via EDM}
{Chiche \MakeLowercase{\textit{et al.}}: 3D vs.\ 2.5D U-Net for MRI Super-Resolution via EDM}

\IEEEaftertitletext{\vspace{-4pt}\begin{center}\begin{minipage}{0.9\textwidth}\footnotesize\centering
\textit{Preprint notice:} This is version~2 of this preprint: the revised manuscript resubmitted to
\textit{IEEE Access} after peer review and currently under review. It supersedes the results of
version~1, which were obtained under a different evaluation protocol. The text is identical to the
resubmitted manuscript. This preprint is made available under IEEE's author-posting policy.
Source code: \url{https://github.com/Magic-Solutions/MRI-Super-Resolution}.
Pretrained weights: \url{https://huggingface.co/Chichonnade/Comparative-Analysis-3D-2.5D-MRI-Super-Resolution-EDM}.
\end{minipage}\end{center}\vspace{6pt}}

\maketitle

\begin{abstract}
Magnetic resonance imaging (MRI) super-resolution (SR) methods that
computationally enhance low-resolution acquisitions offer a software
alternative to higher-field scanners.  We study an elucidated diffusion
model (EDM) framework for $2\times$ in-plane brain MRI SR and compare two
U-Net backbones: (i)~a 3D convolutional U-Net that processes volumetric
patches and (ii)~a 2.5D slice-conditioned U-Net that super-resolves each
slice while conditioning on its two neighbouring slices.  All models
are trained from scratch to convergence on 500 subjects of the NKI
cohort of FOMO60K, with a separate 6-subject validation set for
checkpoint selection, and evaluated on 70 held-out subjects with
subject-level statistics against baselines (EDSR, Swin2SR, Res-SRDiff)
trained on the same data.  Neither architecture is better overall.  On
the degradation used in training the two are tied (\textbf{37.72\,dB}
PSNR for the 3D model and \textbf{37.70\,dB} for the 2.5D model),
0.2\,dB ahead of EDSR.  When the test degradation changes to k-space
truncation, EDSR matches the diffusion models, and with added noise the
2.5D model is the most accurate method.  When the test cohort changes
to 70 subjects of an external dataset (IXI), the 3D model is the most
accurate, 0.4\,dB ahead of the best baseline and 0.9\,dB ahead of the
2.5D model in foreground PSNR.
\end{abstract}

\begin{IEEEkeywords}
Brain MRI, convolutional neural network, denoising diffusion model,
elucidated diffusion model, image super-resolution, 3D U-Net.
\end{IEEEkeywords}

\section{Introduction}
\label{sec:introduction}
\IEEEPARstart{C}{linical} MRI scanners operating at 1.5\,T remain the most
widely installed systems worldwide.  Although 3\,T and 7\,T scanners
deliver higher signal-to-noise ratio and finer spatial resolution, their
substantially higher procurement and maintenance costs restrict
availability, particularly in low-resource settings~\cite{b11}.
Computational super-resolution provides a software-based pathway to
bridge this gap.

Traditional interpolation techniques---bicubic, trilinear---produce
overly smooth outputs that fail to recover fine anatomical detail.
Convolutional neural networks such as SRCNN~\cite{b9} and
EDSR~\cite{b10} improved upon these baselines for natural images.
Three-dimensional convolutional networks have been proposed for brain
MRI SR~\cite{b6,b7}, demonstrating that exploiting inter-slice context
yields measurable gains of 1--3\,dB PSNR.

Denoising diffusion models~\cite{b1} have emerged as powerful
generative frameworks for image synthesis and restoration.
Saharia et al.~\cite{b3} demonstrated with SR3 that conditioning a
diffusion model on a low-resolution input achieves state-of-the-art
perceptual quality for image SR.  Karras et al.~\cite{b18} later
elucidated the design space of diffusion models, introducing a
continuous-sigma formulation with improved conditioning that we adopt
in this work.

Our contributions are as follows:
\begin{enumerate}
  \item We adapt the EDM framework~\cite{b18}, using the U-Net and
        continuous-sigma conditioning codebase of DIAMOND~\cite{b20},
        to volumetric MRI super-resolution, with a 3D variant and a
        2.5D variant conditioned on both neighbouring slices.
  \item We compare them under one matched protocol: the same
        500-subject training set and degradation, training to
        convergence with best-validation checkpoint selection, one
        deterministic sampler with the same number of steps, 70
        held-out test subjects, and subject-level
        statistics with paired tests, against EDSR, Swin2SR and the MRI
        diffusion model Res-SRDiff~\cite{b13}, all three trained from
        scratch on the same data.
  \item We test what a comparison on the training degradation leaves
        open.  The models are trained on block averaging in the image
        domain, whereas a scanner acquires a low-resolution image by
        sampling fewer k-space frequencies, so we evaluate the trained
        models, without retraining, on k-space truncation with and
        without noise, and on an external cohort (IXI).
\end{enumerate}

Under this protocol neither architecture dominates.  On the training
degradation the 3D and 2.5D models are tied and ahead of every
baseline; the 2.5D model is the more robust to a change of degradation
(k-space truncation, noise) and the 3D model to a change of cohort
(IXI).  EDSR trained on the same data matches the diffusion models on
PSNR under k-space truncation while trailing them on perceptual quality
(LPIPS) on every test set.

Throughout, ``super-resolution'' means image-domain $2\times$ in-plane
upsampling of synthetically degraded $128\times128$ slices to
$256\times256$; it recovers detail removed by a simulated degradation
and does not add physical resolution beyond the acquisition.  All data
come from one cohort of healthy volunteers.

\section{Related Work}
\label{sec:related}

\subsection{CNN-Based MRI Super-Resolution}
SRCNN~\cite{b9} was among the first to apply deep learning to SR, and
EDSR~\cite{b10} improved upon it with residual scaling.  For MRI,
Oktay et al.~\cite{b28} trained a CNN on multi-view cardiac
acquisitions; Chen et al.~\cite{b6,b27} proposed 3D densely connected
networks (DCSRN and its GAN variant mDCSRN) for brain MRI, trained on
the Human Connectome Project~\cite{b8}; and Pham et al.~\cite{b7}
extended this to multi-scale 3D architectures.  SMORE~\cite{b29}
trains a through-plane SR network from the in-plane slices of the
same scan, which needs no external HR data.  SynthSR~\cite{b30}
synthesises training pairs from segmentations so that one network
turns heterogeneous clinical scans into 1\,mm T1-weighted volumes, and
validates on downstream morphometry rather than on image metrics
alone.  Transformer SR models such as SwinIR~\cite{b31} and its
successor Swin2SR~\cite{b23} have been applied to MRI by training or
fine-tuning on medical data; we include Swin2SR trained from scratch
on our data.  Our models differ from these in two respects: they are
diffusion models, and they are trained and evaluated under one matched
protocol with the CNN and transformer baselines rather than compared
across papers.

\subsection{Diffusion-Based Super-Resolution}
Score-based and denoising diffusion models~\cite{b1,b2,b14,b15}
reverse a noising process with a learned denoiser.  SR3~\cite{b3}
conditions a DDPM on the LR input via channel concatenation and samples
from pure noise.  Two lines of work reduce the cost of that sampling.
Partial diffusion (Zhao et al.~\cite{b12}) starts the reverse process
from a partially noised LR image rather than from noise, on the
argument that the LR image is already close to the HR one.  Residual
shifting
(ResShift~\cite{b32}; Res-SRDiff~\cite{b13} for MRI) defines the
diffusion over the residual between the LR and HR images so that a
few steps suffice; Res-SRDiff reconstructs an MRI slice in four steps.
DisC-Diff~\cite{b33} conditions a diffusion model on a second
contrast for multi-contrast MRI SR, and score-based models have also
been used for accelerated MRI reconstruction from undersampled
k-space~\cite{b34}.  We train Res-SRDiff on our data as a diffusion
comparator (Section~\ref{sec:baselines}) and compare the three designs
at the algorithm level in Section~\ref{sec:relation}.

\subsection{Elucidated Diffusion Models}
Karras et al.~\cite{b18} reformulated diffusion models using a
continuous noise level $\sigma$ instead of discrete timesteps, with
preconditioning functions $c_\text{in}$, $c_\text{out}$,
$c_\text{skip}$, and $c_\text{noise}$ that stabilize training and
improve sample quality.  This formulation was adopted by
DIAMOND~\cite{b20}, a diffusion world model for video games whose code
includes a conditional EDM upsampler;
our U-Net and training infrastructure come from that code
(Section~\ref{sec:training} lists what was changed).

\section{Methods}
\label{sec:methods}

\subsection{Problem Formulation}

Let $\mathbf{x}_\text{HR} \in \mathbb{R}^{D \times H \times W}$
denote a high-resolution MRI volume and
$\mathbf{x}_\text{LR} \in \mathbb{R}^{D \times H' \times W'}$
its low-resolution counterpart ($H' = H/2$, $W' = W/2$; the depth
dimension~$D$ is preserved, i.e., downsampling is applied in-plane
only).  The SR task seeks a mapping
$f_\theta: \mathbf{x}_\text{LR} \mapsto \hat{\mathbf{x}}_\text{HR}$.

\subsection{EDM Noise Conditioning}

Following Karras et al.~\cite{b18}, we parameterize the diffusion
process using a continuous noise level $\sigma$.  The noisy
observation is:
\begin{equation}\label{eq:noisy}
  \mathbf{x}_\sigma = \mathbf{x}_\text{HR}
    + \sigma \cdot \boldsymbol{\epsilon},
  \quad \boldsymbol{\epsilon} \sim \mathcal{N}(\mathbf{0}, \mathbf{I}).
\end{equation}
The denoiser $D_\theta$ is preconditioned as:
\begin{equation}\label{eq:precond}
  D_\theta(\mathbf{x}_\sigma, \sigma)
  = c_\text{skip}\,\mathbf{x}_\sigma
    + c_\text{out}\,F_\theta\!\bigl(
        c_\text{in}\,\mathbf{x}_\sigma,\;
        c_\text{noise},\;
        \mathbf{x}_\text{LR}
      \bigr),
\end{equation}
where $F_\theta$ is the U-Net backbone, $\sigma_\text{data} = 0.5$ is
the standard deviation assumed for the data, as in EDM~\cite{b18}, and
the scaling functions are:
\begin{align}
  c_\text{in}(\sigma)   &= 1/\sqrt{\sigma^2 + \sigma_\text{data}^2}, \\
  c_\text{skip}(\sigma) &= \sigma_\text{data}^2 / (\sigma^2 + \sigma_\text{data}^2), \\
  c_\text{out}(\sigma)  &= \sigma \cdot \sqrt{c_\text{skip}(\sigma)}, \\
  c_\text{noise}(\sigma)&= \tfrac{1}{4}\ln\sigma.
\end{align}
During training, $\sigma$ is sampled from a log-normal distribution:
$\ln\sigma \sim \mathcal{N}(P_\text{mean}, P_\text{std}^2)$, with the
values of EDM~\cite{b18}, $P_\text{mean} = -1.2$ and $P_\text{std} = 1.2$.
The training loss is:
\begin{equation}\label{eq:loss}
  \mathcal{L} = \mathbb{E}_{\sigma, \mathbf{x}_\text{HR}, \boldsymbol{\epsilon}}
    \bigl[
      \|D_\theta(\mathbf{x}_\sigma, \sigma) - \mathbf{x}_\text{HR}\|^2
    \bigr].
\end{equation}

\subsection{Architecture A: 3D Convolutional U-Net}

The 3D U-Net~\cite{b4,b5} operates on volumetric tensors of shape
$(B, C, D, H, W)$.  The encoder has four levels with channel dimensions
$[32, 64, 128, 256]$, each containing 2~residual blocks with
3D~convolutions ($3\!\times\!3\!\times\!3$), adaptive group normalization
conditioned on $c_\text{noise}$, and SiLU activations.  Multi-head
self-attention~\cite{b16} with flash attention (\texttt{scaled\_dot\_product\_attention})
is applied at the deepest level.  The decoder mirrors the encoder with
skip connections.  Noise conditioning uses Fourier feature embeddings
projected through an MLP.  Total parameters: \textbf{50.7\,M}.

The LR volume is trilinearly upsampled to the target resolution and
concatenated with the noisy HR target along the channel dimension (2
input channels total).

During training, random 3D patches of $64^3$ LR voxels are extracted
together with the matching $64 \times 128 \times 128$ HR voxels (twice
the in-plane size), 8~patches per volume per epoch.  Inference uses sliding-window
processing of $64^3$ LR patches with an overlap of one quarter patch
and linear blending, and the shared sampler of
Section~\ref{sec:sampling}.

\subsection{Architecture B: 2.5D Slice-Conditioned U-Net}
\label{sec:arch_b}

\begin{figure*}[!t]
\centering
\includegraphics[width=\textwidth]{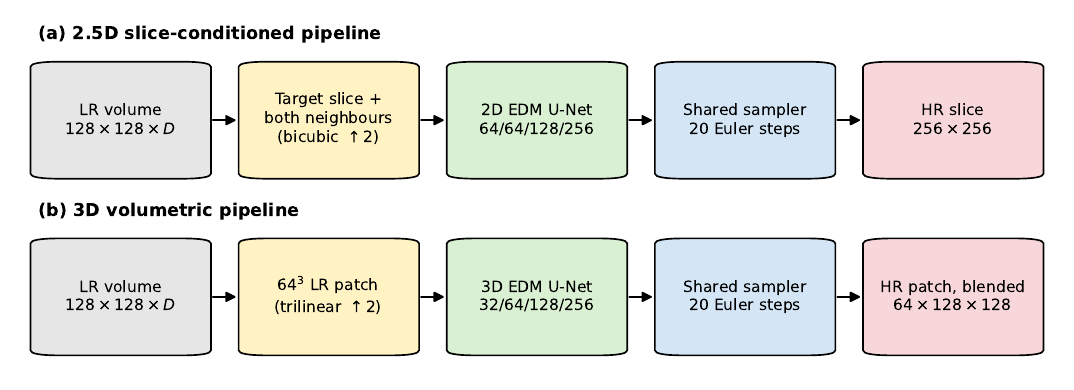}
\caption{Overview of the two super-resolution pipelines.
\textbf{(a)}~The 2.5D model extracts individual slices with one or both
neighbouring slices as context, processes each through a 2D U-Net with
the shared 20-step Euler sampler, and stacks the results.
\textbf{(b)}~The 3D model extracts $64^3$ volumetric patches, processes
them through a 3D U-Net with the same sampler, and blends overlapping
patches to reconstruct the full HR volume.}
\label{fig:method_overview}
\end{figure*}

The 2.5D model decomposes the volumetric SR problem into per-slice
2D~SR tasks with inter-slice conditioning.  For each target slice at
index~$i$ the model receives the target LR slice and its LR
neighbour(s), each bicubic-upsampled to HR resolution, concatenated
along channels with the noisy HR target.  The 2.5D model conditions
on the previous and the next slice (4~input channels); an ablation
variant, \emph{2.5D-prev}, conditions on the previous slice only
(3~input channels).  At the first slice of a volume
the missing previous neighbour is a zero frame, as during training; at
the last slice the missing next neighbour is a repeat of the last
slice, since training never samples a target without a next
neighbour.  The 2D~U-Net has channel dimensions $[64, 64, 128, 256]$
with 2~residual blocks per level, self-attention at the deepest level,
and the same EDM noise conditioning.  Total parameters:
\textbf{51.1\,M} for either variant.  Fig.~\ref{fig:method_overview}
summarises both pipelines.

\subsection{Sampling}
\label{sec:sampling}

Both EDM models are sampled with the same deterministic Euler sampler
(Algorithm~1 of Karras et al.~\cite{b18} without the second-order
correction): 20~steps with noise levels from $\sigma_\text{max} = 5$ to
$\sigma_\text{min} = 2\times10^{-3}$ on the $\rho = 7$ schedule, the
same seed, for both architectures and every test set.
Appendix~\ref{app:sampling} reports 1 to 20 steps; the step count
matters more for the 3D model than for the 2.5D model, so comparing the
architectures at unequal step counts would bias the comparison.

\subsection{Training Configuration}
\label{sec:training}

All models are trained from scratch on the same training set and
degradation (Section~\ref{sec:dataset}) until their validation PSNR
stops improving, and the checkpoint with the best validation PSNR of
each run is the one evaluated on the test set; no checkpoint is chosen
by its test score.  Validation uses the 6 validation subjects and the
sampler of Section~\ref{sec:sampling}.  Appendix~\ref{app:training}
gives the training budget of each model and the stopping rule, and
Section~\ref{sec:convergence} the learning curves.  The EDM
models use AdamW ($\text{lr} = 10^{-4}$, weight decay $10^{-2}$) and the
loss of~(\ref{eq:loss}); the 2.5D models add a 100-step learning-rate
warm-up.  EDSR and Swin2SR use the architectures of~\cite{b10,b23}
with an L1 loss and AdamW ($\text{lr} = 2\times10^{-4}$, no weight decay).  Res-SRDiff uses the
authors' public code and defaults (batch~16, EMA weights,
$4\times$MSE + LPIPS loss).  Learning rates are constant
(no schedule).

\textit{Relation to the DIAMOND code:}
The 2.5D network is the conditional upsampler of DIAMOND's
Counter-Strike configuration~\cite{b20} with its hyperparameters
unchanged: channels $[64, 64, 128, 256]$, two residual blocks per
level, self-attention at the deepest level only, 2048 conditioning
channels, $\sigma_\text{data} = 0.5$, offset noise 0.1, and the
training noise distribution $P_\text{mean} = -1.2$, $P_\text{std} =
1.2$ with $\sigma \in [2\times10^{-3}, 20]$ (DIAMOND's Atari
configuration differs: $[64, 64, 64, 64]$ channels, no attention, 256
conditioning channels, $P_\text{mean} = -0.4$).  Three things were
changed for MRI: one input channel instead of three, an upsampling
factor of 2 instead of 5, and the conditioning on the next slice
(Section~\ref{sec:arch_b}), which adds one input channel.  The
sampler is DIAMOND's deterministic Euler sampler with
$\sigma_\text{max} = 5$ and 20 steps instead of 3.  The 3D network is
a new implementation of the same block structure with 3D convolutions,
narrower channels $[32, 64, 128, 256]$ and 1024 conditioning channels
to fit $64^3$ patches on one NVIDIA L4 GPU (22\,GB).  None of these
settings was tuned for MRI; Section~\ref{sec:steps} and
Appendix~\ref{app:sampling} examine the sampling side, and the
training-side choices (noise distribution, attention placement) are
left as stated.

Algorithm~\ref{alg:training} summarizes the EDM training procedure.

\begin{algorithm}[!t]
\caption{EDM-based MRI Super-Resolution Training}
\label{alg:training}
\begin{algorithmic}[1]
\REPEAT
  \STATE Sample $(\mathbf{x}_\text{LR}, \mathbf{x}_\text{HR})$
         from dataset
  \STATE Sample $\ln\sigma \sim \mathcal{N}(P_\text{mean}, P_\text{std}^2)$
  \STATE $\boldsymbol{\epsilon} \sim \mathcal{N}(\mathbf{0}, \mathbf{I})$
  \STATE $\mathbf{x}_\sigma \leftarrow
         \mathbf{x}_\text{HR} + \sigma \cdot \boldsymbol{\epsilon}$
  \STATE Compute $D_\theta(\mathbf{x}_\sigma, \sigma, \mathbf{x}_\text{LR})$
         via~(\ref{eq:precond})
  \STATE $\mathcal{L} \leftarrow
         \|D_\theta - \mathbf{x}_\text{HR}\|^2$
  \STATE Update $\theta$ via AdamW
\UNTIL{convergence}
\end{algorithmic}
\end{algorithm}

\section{Dataset}
\label{sec:dataset}

We use the NKI (Nathan Kline Institute) cohort of the FOMO60K
dataset~\cite{b19}, which provides T1-weighted structural brain MRI
volumes from over 1{,}300 subjects at approximately 1\,mm isotropic
resolution in NIfTI format.  We keep one session per subject and only
volumes with 176 sagittal slices.

\textit{Ethics and consent:}
The study uses de-identified human brain MRI data from the publicly
available FOMO60K/NKI dataset.  Informed consent was obtained by the
original data collection study, and the present work involves only
secondary analysis of anonymized data.

\textit{Preprocessing and training degradation:}
Each volume is clipped to its 1st--99th percentile range, scaled to
$[0, 255]$ and sliced along the sagittal axis.  We use a fixed
intensity range rather than z-score normalization because PSNR is
defined relative to the intensity range, so a fixed range keeps it
comparable across volumes.  Each $256\times256$ HR slice is downsampled
by a factor of~2 with $2\times2$ block averaging to produce the
$128\times128$ LR input.  The networks receive these intensities
rescaled to $[-1, 1]$; the 3D model is trained on random patches of
the same LR/HR volumes.

\textit{Subject-level split:}
The volumes are split at the subject level into 500 training subjects,
6 validation subjects and 70 test subjects; no subject appears in more
than one split.

\textit{Test degradations:}
The models are trained on block averaging, an image-domain
approximation of low resolution.  A scanner instead acquires a
low-resolution image by sampling only the low spatial frequencies (the
centre of k-space), and its images contain noise.  To test whether
models trained on the approximation still work on a degradation closer
to acquisition, the 70 test volumes are also degraded in two ways that
no learned model saw in training.  \emph{k-space truncation} keeps the
central half of the 2D k-space of each HR slice along each in-plane
axis; unlike block averaging, this leaves ripples next to sharp edges
(Gibbs ringing).  \emph{k-space truncation with noise} additionally
adds complex Gaussian noise of standard deviation 2 (in 0--255
intensity units) to the truncated k-space before the magnitude is
taken, which gives Rician-distributed noise in the LR image.  Both are
simulations from processed magnitude images, not raw multi-coil
acquisitions, and the noise level is not calibrated to a scanner.

\textit{External test set:}
To test generalisation beyond the training cohort, 70 subjects were
drawn at random from the IXI cohort of FOMO60K~\cite{b19,b35} (581
subjects with a T1-weighted scan), acquired at three London hospitals
on Philips 3\,T, Philips 1.5\,T and GE 1.5\,T scanners.  FOMO60K distributes these volumes skull-stripped at
$0.94\times0.94\times1.2$\,mm; we reorient them to RAS, resample them
to 1\,mm isotropic with cubic interpolation, zero-pad each sagittal
slice to $256\times256$, and degrade them by block averaging as the
training data.  Each volume has 241 sagittal slices; slices that contain no brain
are not scored.  No
model was retrained or fine-tuned for this set.

\section{Results}
\label{sec:results}

\subsection{Metrics and Statistics}
\label{sec:metrics}

Every number in this section is produced by one evaluation script.
Per slice we report PSNR, 2-D SSIM~\cite{b17}, LPIPS with AlexNet
features~\cite{b22}, and foreground PSNR (PSNR$_\text{fg}$), the PSNR
over the pixels where the HR slice is non-zero.  We add the last
because the background around the head is empty and every method
reconstructs it almost exactly, which raises plain PSNR without
reflecting anatomy; the effect is largest on the skull-stripped
external test set, where most of each slice is background.  Slices of
one subject are not independent samples, so each metric is averaged
within a test volume and we report the mean and standard deviation over
the 70 subjects; the bootstrap 95\% confidence interval~\cite{b26} of a
mean PSNR is about $\pm0.25$\,dB on the block-averaged test set.
Differences between methods are tested with paired Wilcoxon signed-rank
tests~\cite{b24} over subjects, with Holm correction~\cite{b25}.

\subsection{Baselines}
\label{sec:baselines}

Bilinear and bicubic interpolation; EDSR~\cite{b10} and
Swin2SR~\cite{b23} trained from scratch on our training set and
degradation rather than with their published natural-image weights;
and Res-SRDiff~\cite{b13}, a residual-shifting diffusion model for MRI
SR, trained from scratch on the same data with the authors' code.
Res-SRDiff's training loss includes the LPIPS network that we use as a
metric, so its LPIPS is not comparable with that of the other methods
and is marked ($\dagger$) in the tables.

\subsection{Main Results: Block-Averaged Degradation}

Table~\ref{tab:results} gives the results under the degradation used in
training.  The 3D and 2.5D models are tied at the top: the 3D model is
ahead by 0.02\,dB PSNR, a difference that the paired test detects
($p = 2\times10^{-4}$) but that is far below anything visible, and
SSIM and LPIPS agree to within 0.001.  Both are 0.20--0.23\,dB ahead of
EDSR, the strongest baseline, and 0.27--0.29\,dB ahead of the 2.5D-prev
variant, and both have a lower LPIPS than the CNN and transformer
baselines (0.022--0.023 against 0.026--0.027).  Every learned model
improves on bicubic interpolation by 2.8--4.1\,dB.

\begin{table}[!t]
\centering
\caption{$2\times$ super-resolution on the 70 held-out NKI test
subjects under the degradation used in training (block averaging).
Mean over subjects ($\pm$ SD over subjects for PSNR).  $\dagger$: not
comparable with the other methods (Section~\ref{sec:baselines}).}
\label{tab:results}
\footnotesize
\setlength{\tabcolsep}{3.5pt}
\begin{tabular}{@{}lcccc@{}}
\toprule
\textbf{Method} & \textbf{PSNR$\uparrow$} & \textbf{PSNR$_\text{fg}\uparrow$} & \textbf{SSIM$\uparrow$} & \textbf{LPIPS$\downarrow$} \\
\midrule
Bilinear                   & $31.91 \pm 1.02$ & 29.22 & 0.939 & 0.115 \\
Bicubic                    & $33.66 \pm 1.06$ & 30.91 & 0.957 & 0.087 \\
EDSR~\cite{b10}            & $37.50 \pm 1.11$ & 34.69 & 0.977 & 0.026 \\
Swin2SR~\cite{b23}         & $37.43 \pm 1.10$ & 34.71 & 0.976 & 0.027 \\
Res-SRDiff~\cite{b13}      & $36.48 \pm 1.10$ & 33.80 & 0.972 & 0.008$^\dagger$ \\
2.5D (ours)                & $37.70 \pm 1.08$ & 34.88 & \textbf{0.978} & \textbf{0.022} \\
2.5D-prev (ours)           & $37.43 \pm 1.08$ & 34.61 & 0.977 & 0.023 \\
3D (ours)                  & $\mathbf{37.72 \pm 1.07}$ & \textbf{34.91} & \textbf{0.978} & 0.023 \\
\bottomrule
\end{tabular}
\end{table}

\subsection{Robustness to k-Space Degradation}

Table~\ref{tab:kspace} evaluates the same checkpoints on the two
k-space test sets.  Without noise, every learned model loses
2.3--2.9\,dB relative to Table~\ref{tab:results}, and the best models
gain only about 1\,dB over bicubic interpolation.  EDSR and the 2.5D
model are tied at the top (0.04\,dB apart), Swin2SR and the 3D model
are 0.04 and 0.11\,dB behind the 2.5D model, and the diffusion models
keep a lower LPIPS (0.035--0.039 against 0.041--0.044).  With noise the
ranking changes: the 2.5D model is the best method on every metric,
0.28\,dB ahead of EDSR and 0.30\,dB ahead of the 3D model in PSNR
($p < 10^{-12}$), and the 3D model falls to the level of EDSR.  No
model was trained with noise, so this set measures robustness, not
denoising.

\begin{table*}[!t]
\centering
\caption{Same checkpoints as Table~\ref{tab:results} on the two
k-space test sets (70 subjects; mean over subjects, $\pm$ SD for PSNR).
$\dagger$: not comparable with the other methods
(Section~\ref{sec:baselines}).}
\label{tab:kspace}
\begin{tabular}{@{}lcccccccc@{}}
\toprule
& \multicolumn{4}{c}{\textbf{k-space truncation}} & \multicolumn{4}{c}{\textbf{k-space truncation + Rician noise}} \\
\cmidrule(lr){2-5} \cmidrule(lr){6-9}
\textbf{Method} & PSNR & PSNR$_\text{fg}$ & SSIM & LPIPS & PSNR & PSNR$_\text{fg}$ & SSIM & LPIPS \\
\midrule
Bicubic              & $33.93 \pm 1.06$ & 31.22 & 0.958 & 0.077 & $32.63 \pm 0.67$ & 30.93 & 0.945 & 0.103 \\
EDSR                 & $\mathbf{35.01 \pm 1.04}$ & 32.24 & \textbf{0.962} & 0.041 & $33.26 \pm 0.60$ & 31.67 & 0.942 & 0.081 \\
Swin2SR              & $34.94 \pm 1.06$ & \textbf{32.26} & 0.961 & 0.044 & $33.02 \pm 0.57$ & 31.56 & 0.941 & 0.081 \\
Res-SRDiff           & $34.20 \pm 1.03$ & 31.50 & 0.955 & 0.019$^\dagger$ & $33.00 \pm 0.70$ & 30.76 & 0.935 & 0.058$^\dagger$ \\
2.5D (ours)          & $34.98 \pm 1.01$ & 32.17 & \textbf{0.962} & 0.036 & $\mathbf{33.54 \pm 0.64}$ & \textbf{31.70} & \textbf{0.946} & \textbf{0.061} \\
2.5D-prev (ours)     & $34.80 \pm 1.00$ & 32.00 & 0.961 & \textbf{0.035} & $33.30 \pm 0.61$ & 31.46 & 0.944 & 0.062 \\
3D (ours)            & $34.87 \pm 1.03$ & 32.07 & 0.961 & 0.039 & $33.24 \pm 0.62$ & 31.48 & 0.942 & 0.071 \\
\bottomrule
\end{tabular}
\end{table*}

\subsection{External Validation on IXI}
\label{sec:ixi}

\begin{figure*}[!t]
\centering
\includegraphics[width=\textwidth]{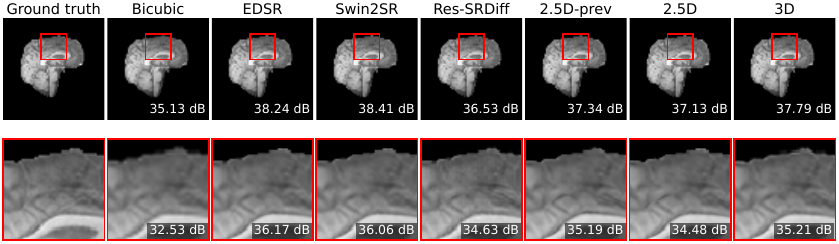}
\caption{All methods on a mid-sagittal slice of IXI subject sub\_10655
(slice 120 of 241), the volume whose 2.5D foreground PSNR is closest to
the 70-subject median; layout as in Fig.~\ref{fig:visual_all}.  The
slice is chosen by this rule, not by appearance: on it EDSR and Swin2SR
score higher than the 3D model, whereas over the 70 subjects the 3D
model leads (Table~\ref{tab:ixi}).  The slice PSNR is inflated by the
empty background of the skull-stripped volume.}
\label{fig:ixi_example}
\end{figure*}

Table~\ref{tab:ixi} applies the same checkpoints, without retraining,
to the 70 IXI subjects.  Because most of each skull-stripped slice is
empty background, the ranking is read from foreground PSNR
(Section~\ref{sec:metrics}).  The ranking differs from NKI.  The 3D
model is the most accurate: 0.36\,dB ahead of Swin2SR, 0.61\,dB ahead
of EDSR and 0.88\,dB ahead of the 2.5D model, on at least 67 of the 70
subjects in each comparison.  The 2.5D model falls below the CNN and
transformer baselines on foreground PSNR but has the lowest LPIPS
(0.0089 against 0.0095 for the 3D model and 0.011 for the baselines).
The gap between the two architectures does not come from the 3D
model's longer training, since its epoch-20 checkpoint is already
0.74\,dB ahead of the 2.5D model.  On the two volumes we inspected it
holds inside the brain as well as along the edge of the skull-stripping
mask, so it is not an artefact of the mask.  Fig.~\ref{fig:ixi_example}
shows one IXI slice.

\begin{table}[!t]
\centering
\caption{Zero-shot evaluation on 70 subjects of the IXI cohort
(skull-stripped, resampled to 1\,mm).  Mean over subjects ($\pm$ SD
over subjects for foreground PSNR, the metric the ranking is read
from).  $\dagger$: not comparable with the other methods
(Section~\ref{sec:baselines}).}
\label{tab:ixi}
\footnotesize
\setlength{\tabcolsep}{3.5pt}
\begin{tabular}{@{}lcccc@{}}
\toprule
\textbf{Method} & \textbf{PSNR$_\text{fg}\uparrow$} & \textbf{PSNR$\uparrow$} & \textbf{SSIM$\uparrow$} & \textbf{LPIPS$\downarrow$} \\
\midrule
Bilinear                   & $28.95 \pm 0.94$ & 37.41 & 0.9798 & 0.0637 \\
Bicubic                    & $31.16 \pm 0.96$ & 39.64 & 0.9898 & 0.0446 \\
EDSR~\cite{b10}            & $33.04 \pm 0.93$ & 41.89 & 0.9944 & 0.0107 \\
Swin2SR~\cite{b23}         & $33.29 \pm 0.97$ & 41.65 & 0.9944 & 0.0115 \\
Res-SRDiff~\cite{b13}      & $31.54 \pm 1.02$ & 40.09 & 0.9918 & 0.0093$^\dagger$ \\
2.5D (ours)                & $32.77 \pm 1.00$ & 41.64 & 0.9942 & \textbf{0.0089} \\
2.5D-prev (ours)           & $32.54 \pm 1.09$ & 41.40 & 0.9938 & 0.0102 \\
3D (ours)                  & $\mathbf{33.65 \pm 0.97}$ & \textbf{42.29} & \textbf{0.9949} & 0.0095 \\
\bottomrule
\end{tabular}
\end{table}

\subsection{Number of Sampling Steps}
\label{sec:steps}

Five Euler steps reproduce the 20-step quality of both architectures
(within 0.03\,dB), whereas a single step costs the 2.5D design 0.6\,dB
and the 3D model 4.4\,dB.  Few steps suffice because at high noise
levels the denoiser already returns its estimate of the HR slice from
the LR conditioning.  Beyond two to five steps the PSNR is flat: it
changes by at most 0.13\,dB up to 20 steps, and more steps do not
improve it.  Appendix~\ref{app:sampling} gives the step sweep, the
second-order solver and the noise-level analysis.

\subsection{Convergence}
\label{sec:convergence}

\begin{figure}[!htbp]
\centering
\includegraphics[width=\columnwidth]{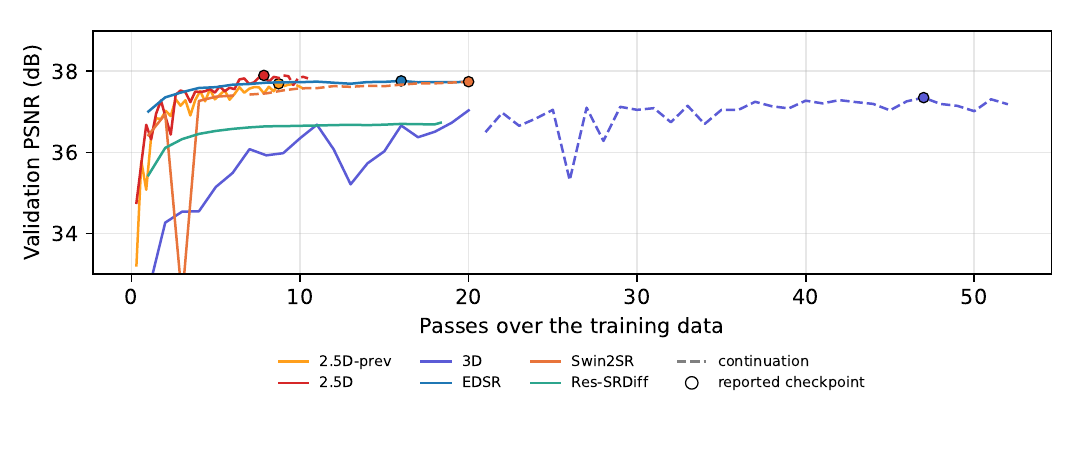}
\caption{Validation PSNR of every run against passes over the training
data (2.5D epochs are 400 steps; Res-SRDiff validates every 5{,}500
steps).  Solid: initial budget; dashed: continuation with early
stopping; circles: the reported checkpoints.  The validation PSNR is
sampled on 16 slices per subject.}
\label{fig:curves}
\end{figure}

Validation PSNR after every epoch is shown in
Fig.~\ref{fig:curves} for all six runs, initial budget and
continuation.  EDSR and Res-SRDiff plateaued within their initial
budgets (change over the last quarter $+0.00$ and $+0.06$\,dB; best
epochs 16 of 20 and the last evaluation).  The 2.5D-prev run gained
$+0.08$\,dB over its last quarter and its continuation found no better
epoch, so epoch 30 is reported.  The 2.5D run's best validation epoch
is 27 (37.90\,dB); its continuation from epoch 30 came within
0.001\,dB of that at epoch 31 and then stopped, so epoch 27 is
reported and the run is treated as converged within its initial budget.  The 3D run, which was still rising at
the end of its initial budget ($+1.01$\,dB over its last quarter),
gained a further $+0.31$\,dB on validation over 27 continuation epochs
(best at epoch 47, stopped after 52); its reported checkpoint is epoch
47.  Swin2SR gained $+0.34$\,dB on validation over 14 continuation
epochs and was still improving by about 0.02\,dB per epoch when it
reached its 20-epoch cap; its epoch-20 checkpoint is reported, and it is
the one model whose convergence is not established by the stopping
rule.  The validation PSNR is sampled (16 slices per subject), so
single-epoch swings of 0.3--0.5\,dB, visible for the 3D run, are
noise.

\subsection{Qualitative Results}
\label{sec:qualitative}

\begin{figure*}[!t]
\centering
\includegraphics[width=\textwidth]{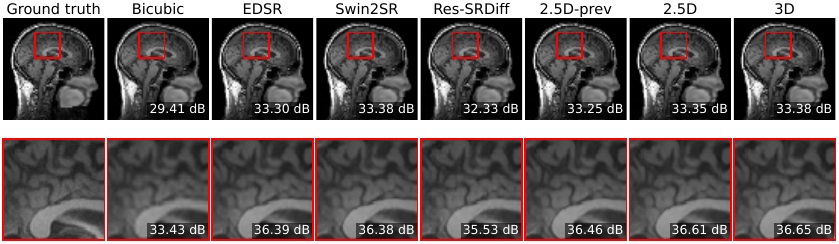}
\caption{All methods on a mid-sagittal slice of test volume sub\_133
(slice 88 of 176; the volume whose 2.5D PSNR is closest to the 70-volume
median).  Top: full slice with the slice PSNR; bottom: the red region
with the PSNR of the crop.}
\label{fig:visual_all}
\end{figure*}

\begin{figure}[!htbp]
\centering
\includegraphics[width=\columnwidth]{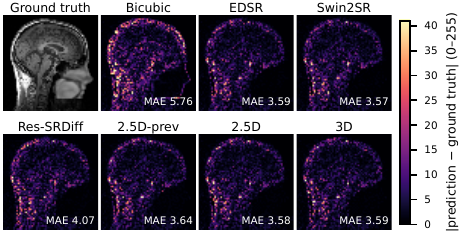}
\caption{Absolute error $|\hat{x} - x|$ of the slice of
Fig.~\ref{fig:visual_all} in 0--255 intensity units, one colour scale
for every method, with the mean absolute error over the foreground.
Every learned method leaves its largest errors at the skull and at
gray--white boundaries; bicubic interpolation adds errors along the
whole cortical ribbon.}
\label{fig:error_heatmap}
\end{figure}

\begin{figure*}[!t]
\centering
\includegraphics[width=\textwidth]{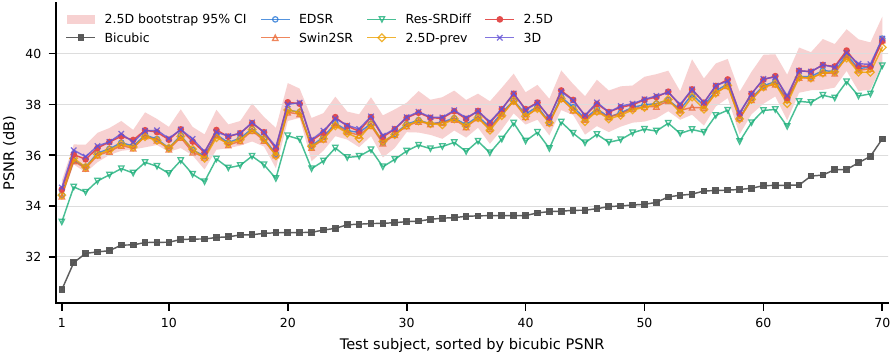}
\caption{Per-subject PSNR of every method on the 70 block-averaged
test volumes, sorted by bicubic PSNR.  The band is the 2.5D model's
bootstrap 95\% confidence interval over the slices of each subject
(10{,}000 resamples); it reflects the within-subject spread between
lateral and mid-sagittal slices, not the between-subject interval of
Table~\ref{tab:results}.}
\label{fig:per_subject}
\end{figure*}

Fig.~\ref{fig:visual_all} compares all methods on the mid-sagittal
slice of the test volume whose 2.5D PSNR is closest to the median
(sub\_133, slice 88), with a zoomed region over the cortex and the
corpus callosum, and Fig.~\ref{fig:error_heatmap} shows the absolute
error of the same slice on one colour scale.  Apart from Res-SRDiff,
the learned methods are within 0.13\,dB of each other on this slice
(33.25--33.38\,dB, against 29.41\,dB for bicubic interpolation) and are
visually close; bicubic interpolation blurs the cortical ribbon, and
every learned method concentrates its remaining error at the scalp, the
skull and the gray--white boundaries.  Fig.~\ref{fig:per_subject} gives the
per-subject PSNR of every method over the 70 test volumes: the five
best methods (3D, 2.5D, EDSR, 2.5D-prev, Swin2SR) move together from
subject to subject and lie within 0.7\,dB of each other on every subject
(median 0.3\,dB), and Res-SRDiff is below all five on all 70.

\begin{figure*}[!t]
\centering
\includegraphics[width=\textwidth]{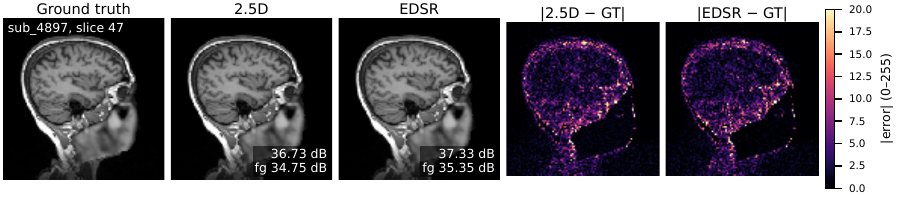}
\caption{A failure case: the slice with substantial anatomy
(foreground fraction 0.63) where the 2.5D model loses most foreground
PSNR against EDSR (sub\_4897, slice 47; $-0.6$\,dB).  Left to right:
ground truth, the two reconstructions with slice and foreground PSNR,
and their absolute errors on one scale.}
\label{fig:failure}
\end{figure*}

Fig.~\ref{fig:failure} shows where the 2.5D model loses most against
EDSR.  Over all 12{,}310 scored slices the largest foreground-PSNR
losses are at the first and last slices of a volume, where one of the
conditioning neighbours is the edge padding (mean 2.5D $-$ EDSR
foreground PSNR: $-0.24$\,dB on slices 0--5 and $-0.27$\,dB on slices
171--175, against $+0.31$\,dB on slices 21--80); the 20 worst slices
all lie within ten slices of a volume edge.  Among slices with
substantial anatomy the largest loss is 0.6\,dB (sub\_4897, slice 47,
shown): most of the 2.5D model's extra error lies along the scalp and
the skull, and inside the brain the two error maps are similar.
sub\_4897 is also the only test volume where the 2.5D model's
per-volume PSNR is below EDSR's ($-0.09$\,dB).

\section{Discussion}
\label{sec:discussion}

\subsection{3D versus 2.5D}
Under a matched protocol neither architecture is uniformly better.
On the training degradation the 3D model and the 2.5D model are tied
(37.72 and 37.70\,dB).  A change of degradation favours the 2.5D model:
it is 0.11\,dB ahead of the 3D model under k-space truncation and
0.30\,dB ahead with noise.  A change of cohort favours the 3D model: on
IXI it is 0.88\,dB ahead in foreground PSNR on every one of the 70
subjects, and its initial-budget checkpoint is already 0.74\,dB ahead,
so the difference does not come from training length.  For the 2.5D
design, conditioning on both neighbouring slices matters: it adds
0.18--0.27\,dB over the previous-slice-only variant on every test set.
Why the 3D model transfers better to IXI is not established here.  It
is not the edge of the skull-stripping mask: on the volumes we
inspected the gap is at least as large inside the brain.  Two untested
candidates remain: the 3D model's larger spatial context ($64^3$
patches against three slices), and the different scanners and
preprocessing of IXI (resampling from 1.2\,mm), which change the image
content the 2.5D model conditions on.  The comparison is sensitive to the protocol: the
step sweep (Table~\ref{tab:steps}) shows that sampling the 2.5D model
with one step and the 3D model with twenty would put the 3D model
0.5\,dB ahead, and unequal passes over the data or a test set of a few
subjects would move the result by similar amounts.  With equal
sampling, equal data and 70 test subjects the two architectures are
within 0.03\,dB on the training cohort.  The 2.5D model does so at a
third of the 3D model's inference cost.

\subsection{Diffusion Models versus Trained CNN Baselines}
EDSR trained on our data comes within 0.2\,dB of the best diffusion
models on the training degradation and matches them under k-space
truncation, with 1.4\,M parameters and 0.8\,s per volume; on IXI the
CNN and transformer baselines transfer better than the 2.5D model but
not as well as the 3D model.  The
diffusion models keep two advantages: lower LPIPS on every test set
(0.022 against 0.026 for EDSR on block averaging; 0.035 against 0.041
on k-space; 0.009 against 0.011 on IXI), and robustness to noise, where
the 2.5D model leads every CNN and transformer baseline on every
metric.  Res-SRDiff,
trained with the authors' code and loss, is 1.2\,dB below the EDM
models in PSNR on the training degradation; its very low LPIPS is a
consequence of training on that metric and is reported for
completeness only.

\subsection{Relation to Other Diffusion SR Designs}
\label{sec:relation}
The three diffusion designs in this study differ in what the reverse
process starts from and what it predicts.  Our EDM models start from
Gaussian noise at $\sigma_\text{max}$, predict the clean HR slice at
every noise level with a denoiser conditioned on the LR slices, and
reach their accuracy in five deterministic steps
(Table~\ref{tab:steps}).  Partial diffusion~\cite{b12} starts instead
from the LR image noised to an intermediate level, which saves the
early steps but ties the result to the LR image where the degradation
is strong; residual shifting~\cite{b13,b32} moves the diffusion to the
LR--HR residual with a schedule designed for a handful of steps.
Res-SRDiff trained on our data is 1.0--1.2\,dB below the EDM models on
the training degradation and 0.6--0.8\,dB below on k-space truncation, with
four steps against five for the same accuracy from the EDM models;
its loss includes LPIPS-Alex, so its perceptual scores are not
comparable.  The step sweep suggests that for $2\times$ in-plane SR
the number of steps is not the bottleneck of the EDM design, which
weakens the case for the shortcuts the other two designs take, at
least at this scale factor.

\subsection{Degradation Model}
Block averaging is the degradation the models were trained on, and it
is where every learned method gains most over interpolation
(2.8--4.1\,dB).  k-space truncation is closer to how a lower
acquisition matrix arises and produces Gibbs ringing that block
averaging does not; on it the same models gain only about 1\,dB over
bicubic interpolation, and the ranking among the top models is within
a few hundredths of a dB.  The results in Table~\ref{tab:results}
should therefore be read as an upper bound for this family of models
on matched synthetic degradation, and Table~\ref{tab:kspace} as the
more conservative estimate.  Both remain simulations from processed
magnitude images.

\subsection{Inference Efficiency}
With the shared sampler at 20 steps, one A100 processes a 176-slice
volume in 0.8\,s with EDSR, 12\,s with Swin2SR, 27\,s with Res-SRDiff
(four steps), 111\,s with the 2.5D models and 350\,s with the 3D
model; peak GPU memory is 0.5\,GB for the 2.5D and 2.7\,GB for the 3D
model.  Five steps give the same accuracy at a quarter of the time
(28\,s and 88\,s), and one step is acceptable only for the 2.5D model.
The 2.5D models thus fit an interactive setting on a single GPU; the 3D
model remains an offline method.

\subsection{Flash Attention}
We replace the standard quadratic self-attention with PyTorch's
\texttt{scaled\_dot\_product\_attention}, which dispatches to flash
attention on CUDA GPUs when a compatible kernel is available.  This
avoids materializing the full attention matrix and enables batch
size~4 training on 22\,GB L4 GPUs where the original implementation
would require batch size~2.

\subsection{Limitations}
All training uses one cohort (NKI) and one contrast (T1).  The
external test uses one further cohort (IXI), preprocessed differently
from the training data (skull stripped and resampled), and shows that
the ranking of the architectures can change across cohorts; more
cohorts are needed to say which ranking is typical.
The degradations are simulated: in-plane $2\times$ block averaging and
k-space truncation applied to processed 8-bit magnitude images, without
raw multi-coil k-space, parallel imaging, motion, or a scanner-calibrated
noise level, and no model was trained with noise.  Through-plane
super-resolution, the more common clinical need, is not addressed.
Each model was trained once; the subject-level tests quantify
evaluation uncertainty but not training-seed variance.  All subjects
are healthy volunteers, so the behaviour of the models on pathology
(lesions, tumours, atrophy) is untested.  LPIPS is computed with AlexNet
features, which disadvantages no method except that Res-SRDiff trains
on it.  Finally, we report image-quality metrics and robustness to two
simulated degradations only.  Validation practice for MRI SR
increasingly asks for downstream evidence, such as segmentation or
morphometry agreement with the reference scan~\cite{b30} or reader
studies; neither is included here, so no claim about clinical utility
is made.

\section{Conclusion}
\label{sec:conclusion}

We compared a 3D convolutional and a 2.5D slice-conditioned U-Net for
EDM-based brain MRI super-resolution under one matched protocol: the
same 500-subject training set and degradation, training to convergence
with best-validation checkpoints, one deterministic sampler, 70
held-out test subjects and subject-level statistics, with EDSR, Swin2SR
and Res-SRDiff trained on the same data, and tested on three
degradations of the training cohort and on an external cohort.
Neither architecture is uniformly better.  On the training degradation
the 3D model and the 2.5D model, conditioned on both neighbouring
slices, are tied at 37.72 and 37.70\,dB PSNR, ahead of every baseline.
The 2.5D model is the more robust to a change of degradation and leads
on noisy k-space data; the 3D model is the more robust to a change of
cohort and leads on IXI by 0.4\,dB foreground PSNR over the best
baseline.  EDSR trained on the same data matches the diffusion models
on PSNR under k-space truncation but trails them on LPIPS on every test
set.  Five sampling steps suffice, which makes the 2.5D model a
28-second and the 3D model an 88-second per-volume method on one GPU.  These
are results for image-domain $2\times$ in-plane super-resolution of
synthetically degraded slices from one cohort of healthy volunteers;
they do not establish a gain in physical resolution or performance on
pathology.

Future directions include: (1)~explaining why the ranking changes
across cohorts, and validation on further cohorts such as the Human
Connectome Project~\cite{b8} and on other contrasts; (2)~through-plane super-resolution and training
across the three anatomical axes; (3)~degradations derived from raw
k-space and prospective low-field acquisitions; (4)~a downstream
evaluation such as segmentation agreement; and (5)~training-seed
repeats for the two headline models.

\appendices

\section{Training Budgets and Stopping Rule}
\label{app:training}

Table~\ref{tab:budgets} lists the training of each model.  All runs use
one NVIDIA L4 GPU (22\,GB), except Res-SRDiff, which uses one A100.
Each run is trained for an initial budget and then continued from its
final weights with a re-initialised optimizer, with early stopping
after five validations without a new best (three for Swin2SR).  Every
run is validated after each epoch (Res-SRDiff every 5{,}500 steps) on
the 6 validation subjects, using 16 slices per subject (the 3D model:
the 6 full volumes).  An ``epoch'' is not comparable across models: for
the 2.5D models it is a fixed 400~optimizer steps, for the others a
pass over the training data, so the table gives optimizer steps and
wall-clock time.

\begin{table}[!t]
\centering
\caption{Training of the reported models.  Batch = samples per step
$\times$ gradient-accumulation steps (seg.\ = 2.5D training segments);
the 3D model sees 8 patches per volume per pass.  Initial = the
initial budget in optimizer steps (passes over the data in
parentheses); Best ep.\ = the epoch of the reported checkpoint, with the
number of further epochs trained without improvement before early
stopping; Wall-clock = Vertex AI time of the initial run on one GPU,
including data staging.  $^{c}$: stopped at its 20-epoch cap while
still improving by about 0.02\,dB per epoch on validation.}
\label{tab:budgets}
\scriptsize
\setlength{\tabcolsep}{2.5pt}
\begin{tabular}{@{}lrlrll@{}}
\toprule
\textbf{Model} & \textbf{Params} & \textbf{Batch} & \textbf{Initial} & \textbf{Best ep.} & \textbf{Wall-clock} \\
\midrule
2.5D        & 51.1\,M & $4\times8$ seg. & 12{,}000 ($\sim$8.7) & 27 (+9) & 22\,h (L4) \\
2.5D-prev   & 51.1\,M & $4\times8$ seg. & 12{,}000 ($\sim$8.7) & 30 (+5) & 22\,h (L4) \\
3D          & 50.7\,M & $2\times4$ patches & 10{,}000 (20) & 47 (+5) & 36\,h (L4) \\
EDSR        & 1.4\,M  & $4\times4$ slices & 110{,}000 (20) & 16 (+4) & 8\,h (L4) \\
Swin2SR     & 1.0\,M  & $4\times4$ slices & 33{,}000 (6) & 20$^{c}$ & 23\,h (L4) \\
Res-SRDiff  & 56.7\,M & 16 slices & 100{,}881 ($\sim$18) & last & 12\,h (A100) \\
\bottomrule
\end{tabular}
\end{table}

\section{Sampling Steps, Solver and Noise Levels}
\label{app:sampling}

Table~\ref{tab:steps} varies the number of Euler steps of the shared
sampler for the 2.5D-prev and 3D models on the block-averaged test set.
The sweep was run with the 2.5D-prev checkpoint and the 3D model's
initial-budget checkpoint (epoch 20); it characterises the sampler
rather than the final scores.  Inference time is measured for one
176-slice volume on one A100 GPU; it scales with the number of steps
and, for the 3D model, with the patch count.

\begin{table}[!t]
\centering
\caption{Effect of the number of sampling steps (block-averaged test
set, 70 subjects, mean over subjects).  s/vol.: inference time per volume in
seconds.}
\label{tab:steps}
\begin{tabular}{@{}lcccccc@{}}
\toprule
& \multicolumn{3}{c}{\textbf{2.5D-prev}} & \multicolumn{3}{c}{\textbf{3D (ep.\ 20)}} \\
\cmidrule(lr){2-4} \cmidrule(lr){5-7}
\textbf{Steps} & PSNR & PSNR$_\text{fg}$ & s/vol. & PSNR & PSNR$_\text{fg}$ & s/vol. \\
\midrule
1  & 36.85 & 33.96 & 6   & 32.98 & 30.08 & 18 \\
5  & 37.40 & 34.57 & 28  & 37.39 & 34.63 & 88 \\
20 & 37.43 & 34.61 & 111 & 37.37 & 34.61 & 350 \\
\bottomrule
\end{tabular}
\end{table}

The remaining choices are examined on the 6 validation subjects (16
slices each are used for model selection; here all 1{,}056 slices are
scored), so that nothing is tuned on the test set.

\textit{Steps and solver.}  Table~\ref{tab:ablation_steps} extends the
step sweep to 1, 2, 5, 10 and 20 Euler steps and adds the second-order
Heun solver at 20 steps for the 2.5D models.  Both 2.5D models reach
their 20-step accuracy at two steps (within 0.1\,dB; for 2.5D-prev two
steps is the best point), and the 3D model at five.  Heun changes the
20-step result by at most 0.03\,dB at twice the network evaluations.
The 3D sampler has no second-order option, so the solver comparison is
for the 2.5D models only.

\begin{table}[!t]
\centering
\caption{Sampling steps and solver on the 6 validation subjects
(per-volume mean PSNR in dB; s/vol.: time per volume in seconds).}
\label{tab:ablation_steps}
\footnotesize
\setlength{\tabcolsep}{4pt}
\begin{tabular}{@{}lccccccc@{}}
\toprule
& \multicolumn{5}{c}{\textbf{Euler steps}} & \multicolumn{2}{c}{\textbf{Heun, 20 steps}} \\
\cmidrule(lr){2-6} \cmidrule(lr){7-8}
\textbf{Model} & 1 & 2 & 5 & 10 & 20 & PSNR & s/vol. \\
\midrule
2.5D       & 36.68 & 37.32 & 37.30 & 37.34 & 37.31 & 37.28 & 224 \\
2.5D-prev  & 36.46 & 37.17 & 37.04 & 37.09 & 37.07 & 37.04 & 223 \\
3D (ep.\ 20) & 32.52 & 36.70 & 37.06 & 37.06 & 37.04 & --- & --- \\
\bottomrule
\end{tabular}
\end{table}

\begin{figure}[!htbp]
\centering
\includegraphics[width=\columnwidth]{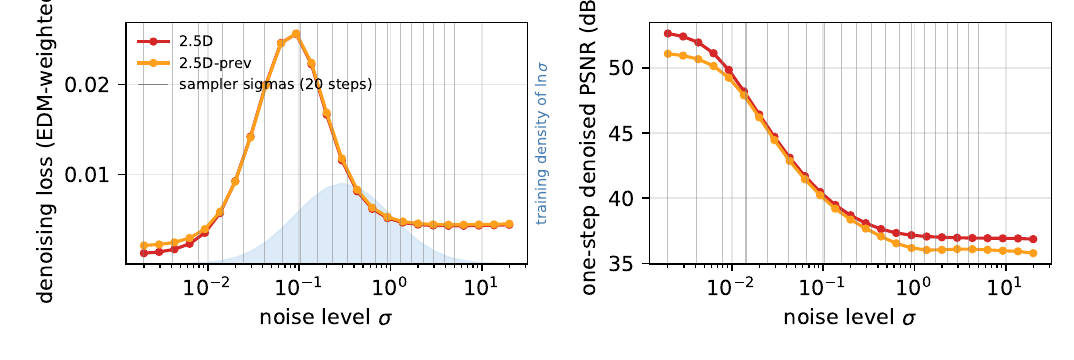}
\caption{Denoising error of the two 2.5D models as a function of the
noise level on the validation volumes.  Left: EDM-weighted denoising
loss, with the training density of $\ln\sigma$ (shaded) and the 20
sampler sigmas (vertical lines).  Right: PSNR of the single-step
denoised estimate at each noise level.}
\label{fig:per_sigma}
\end{figure}

\textit{Noise levels.}  Fig.~\ref{fig:per_sigma} replays the training
forward pass of the two 2.5D models at fixed noise levels.  The
EDM-weighted denoising loss peaks at $\sigma \approx 0.1$, where noise
and anatomical detail have comparable amplitude, and is flat above
$\sigma \approx 2$; the training distribution of $\ln\sigma$
($P_\text{mean} = -1.2$, $P_\text{std} = 1.2$) peaks at $\sigma = 0.3$
and covers every sigma the 20-step sampler visits.  The PSNR of the
single-step denoised estimate falls from 52\,dB at $\sigma = 0.002$ to
the level of the sampled output (37\,dB for the 2.5D model, 36\,dB for
2.5D-prev) for $\sigma \geq 1$: at high noise the denoiser already
returns its estimate of the HR slice from the LR conditioning alone,
which is why two sampling steps suffice.  The two models differ only
at the lowest noise levels, where the extra neighbour adds about 1\,dB.

\begin{thebibliography}{30}

\bibitem{b1}
J.~Ho, A.~Jain, and P.~Abbeel, ``Denoising diffusion probabilistic
models,'' in \emph{Advances in Neural Information Processing Systems},
vol.~33, pp. 6840--6851, 2020.

\bibitem{b2}
A.~Q.~Nichol and P.~Dhariwal, ``Improved denoising diffusion
probabilistic models,'' in \emph{Proc. 38th Int. Conf. Machine
Learning (ICML)}, pp. 8162--8171, 2021.

\bibitem{b3}
C.~Saharia \emph{et al.}, ``Image super-resolution via iterative
refinement,'' \emph{IEEE Trans. Pattern Anal. Mach. Intell.},
vol.~45, no.~4, pp. 4713--4726, 2022.

\bibitem{b4}
O.~Ronneberger, P.~Fischer, and T.~Brox, ``U-Net: Convolutional
networks for biomedical image segmentation,'' in \emph{Medical Image
Computing and Computer-Assisted Intervention -- MICCAI 2015},
pp. 234--241, 2015.

\bibitem{b5}
{\"O}.~{\c{C}}i{\c{c}}ek \emph{et al.}, ``3D U-Net: Learning dense
volumetric segmentation from sparse annotation,'' in \emph{MICCAI
2016}, pp. 424--432, 2016.

\bibitem{b6}
Y.~Chen \emph{et al.}, ``Brain MRI super resolution using 3D deep
densely connected neural networks,'' in \emph{2018 IEEE 15th Int.
Symp. Biomedical Imaging (ISBI)}, pp. 739--742, 2018.

\bibitem{b7}
C.-H.~Pham, A.~Ducournau, R.~Fablet, and F.~Rousseau, ``Multiscale
brain MRI super-resolution using deep 3D convolutional networks,''
\emph{Comput. Med. Imaging Graph.}, vol.~77, p. 101647, 2019.

\bibitem{b8}
D.~C.~Van~Essen \emph{et al.}, ``The WU-Minn Human Connectome
Project: An overview,'' \emph{NeuroImage}, vol.~80, pp. 62--79, 2013.

\bibitem{b9}
C.~Dong, C.~C.~Loy, K.~He, and X.~Tang, ``Image super-resolution
using deep convolutional networks,'' \emph{IEEE Trans. Pattern Anal.
Mach. Intell.}, vol.~38, no.~2, pp. 295--307, 2016.

\bibitem{b10}
B.~Lim, S.~Son, H.~Kim, S.~Nah, and K.~M.~Lee, ``Enhanced deep
residual networks for single image super-resolution,'' in \emph{Proc.
IEEE Conf. Comput. Vis. Pattern Recognit. Workshops}, pp. 136--144,
2017.

\bibitem{b11}
S.~R.~Jannat, K.~Lynch, M.~Fotouhi, S.~Cen, J.~Choupan,
N.~Sheikh-Bahaei, G.~Pandey, and B.~A.~Varghese, ``Advancing 1.5T MR
imaging: Toward achieving 3T quality through deep learning
super-resolution techniques,'' \emph{Frontiers in Human Neuroscience},
vol.~19, p. 1532395, 2025.

\bibitem{b12}
K.~Zhao, K.~Pang, A.~L.~Y.~Hung, H.~Zheng, R.~Yan, and K.~Sung,
``MRI super-resolution with partial diffusion models,'' \emph{IEEE
Trans. Med. Imaging}, vol.~44, no.~3, pp. 1194--1205, 2025, doi:
10.1109/TMI.2024.3483109.

\bibitem{b13}
M.~Safari, S.~Wang, Z.~Eidex, Q.~Li, E.~H.~Middlebrooks, D.~S.~Yu,
and X.~Yang, ``MRI super-resolution reconstruction using efficient
diffusion probabilistic model with residual shifting,'' \emph{arXiv
preprint arXiv:2503.01576}, 2025.

\bibitem{b14}
Y.~Song and S.~Ermon, ``Generative modeling by estimating gradients
of the data distribution,'' in \emph{Advances in Neural Information
Processing Systems}, vol.~32, pp. 11918--11930, 2019.

\bibitem{b15}
P.~Dhariwal and A.~Nichol, ``Diffusion models beat GANs on image
synthesis,'' in \emph{Advances in Neural Information Processing
Systems}, vol.~34, pp. 8780--8794, 2021.

\bibitem{b16}
A.~Vaswani \emph{et al.}, ``Attention is all you need,'' in
\emph{Advances in Neural Information Processing Systems}, vol.~30,
pp. 5998--6008, 2017.

\bibitem{b17}
Z.~Wang, A.~C.~Bovik, H.~R.~Sheikh, and E.~P.~Simoncelli, ``Image
quality assessment: From error visibility to structural similarity,''
\emph{IEEE Trans. Image Processing}, vol.~13, no.~4, pp. 600--612,
2004.

\bibitem{b18}
T.~Karras, M.~Aittala, T.~Aila, and S.~Laine, ``Elucidating the
design space of diffusion-based generative models,'' in \emph{Advances
in Neural Information Processing Systems}, vol.~35,
pp. 26565--26577, 2022.

\bibitem{b19}
S.~Cerri \emph{et al.}, ``A large-scale heterogeneous 3D magnetic
resonance brain imaging dataset for self-supervised learning,''
\emph{arXiv preprint arXiv:2506.14432}, 2025. Dataset (FOMO60K):
\url{https://huggingface.co/datasets/FOMO-MRI/FOMO60K} (NKI cohort
under the Rockland Sample data usage agreement).

\bibitem{b20}
E.~Alonso \emph{et al.}, ``DIAMOND: Diffusion for world modeling:
Visual details matter in Atari,'' in \emph{Advances in Neural
Information Processing Systems}, vol.~37, 2024.

\bibitem{b22}
R.~Zhang, P.~Isola, A.~A.~Efros, E.~Shechtman, and O.~Wang,
``The unreasonable effectiveness of deep features as a perceptual
metric,'' in \emph{Proc. IEEE Conf. Comput. Vis. Pattern Recognit.
(CVPR)}, 2018, pp.~586--595.

\bibitem{b23}
M.~V.~Conde, U.-J.~Choi, M.~Burchi, and R.~Timofte,
``Swin2SR: SwinV2 Transformer for compressed image
super-resolution and restoration,'' in \emph{European Conf. Comput.
Vis. Workshops (ECCVW)}, 2022, pp.~669--687.

\bibitem{b24}
F.~Wilcoxon, ``Individual comparisons by ranking methods,''
\emph{Biometrics Bulletin}, vol.~1, no.~6, pp. 80--83, 1945.

\bibitem{b25}
S.~Holm, ``A simple sequentially rejective multiple test procedure,''
\emph{Scandinavian Journal of Statistics}, vol.~6, no.~2, pp. 65--70,
1979.

\bibitem{b26}
B.~Efron and R.~J.~Tibshirani, \emph{An Introduction to the Bootstrap}.
New York, NY, USA: Chapman \& Hall, 1993.

\bibitem{b27}
Y.~Chen, F.~Shi, A.~G.~Christodoulou, Y.~Xie, Z.~Zhou, and D.~Li,
``Efficient and accurate MRI super-resolution using a generative
adversarial network and 3D multi-level densely connected network,'' in
\emph{Medical Image Computing and Computer Assisted Intervention --
MICCAI 2018}, pp. 91--99, 2018.

\bibitem{b28}
O.~Oktay \emph{et al.}, ``Multi-input cardiac image super-resolution
using convolutional neural networks,'' in \emph{Medical Image Computing
and Computer-Assisted Intervention -- MICCAI 2016}, pp. 246--254, 2016.

\bibitem{b29}
C.~Zhao, B.~E.~Dewey, D.~L.~Pham, P.~A.~Calabresi, D.~S.~Reich, and
J.~L.~Prince, ``SMORE: A self-supervised anti-aliasing and
super-resolution algorithm for MRI using deep learning,'' \emph{IEEE
Trans. Med. Imaging}, vol.~40, no.~3, pp. 805--817, 2021.

\bibitem{b30}
J.~E.~Iglesias \emph{et al.}, ``SynthSR: A public AI tool to turn
heterogeneous clinical brain scans into high-resolution T1-weighted
images for 3D morphometry,'' \emph{Science Advances}, vol.~9, no.~5,
eadd3607, 2023.

\bibitem{b31}
J.~Liang, J.~Cao, G.~Sun, K.~Zhang, L.~Van~Gool, and R.~Timofte,
``SwinIR: Image restoration using Swin Transformer,'' in \emph{Proc.
IEEE/CVF Int. Conf. Comput. Vis. Workshops (ICCVW)}, 2021,
pp. 1833--1844.

\bibitem{b32}
Z.~Yue, J.~Wang, and C.~C.~Loy, ``ResShift: Efficient diffusion model
for image super-resolution by residual shifting,'' in \emph{Advances in
Neural Information Processing Systems}, vol.~36, 2023.

\bibitem{b33}
Y.~Mao, L.~Jiang, X.~Chen, and C.~Li, ``DisC-Diff: Disentangled
conditional diffusion model for multi-contrast MRI super-resolution,''
in \emph{Medical Image Computing and Computer Assisted Intervention --
MICCAI 2023}, pp. 387--397, 2023.

\bibitem{b34}
H.~Chung and J.~C.~Ye, ``Score-based diffusion models for accelerated
MRI,'' \emph{Medical Image Analysis}, vol.~80, 102479, 2022.

\bibitem{b35}
IXI -- Information eXtraction from Images (EPSRC GR/S21533/02). IXI
Dataset. [Online]. Available: \url{https://brain-development.org/ixi-dataset/}

\end{thebibliography}
\end{document}